\documentclass{article}

\usepackage{amsmath} 
\usepackage{amsfonts} 

\newcommand{\matt}[1]{} 

\usepackage{amsthm}

\usepackage{fontenc} 

\usepackage{textcomp} 

\usepackage{placeins}

\usepackage{times}
\usepackage{graphicx} 

\usepackage{natbib}

\usepackage{algorithm}
\usepackage{algorithmic}

\usepackage{hyperref}

\usepackage[accepted]{icml2017}

\icmltitlerunning{Binarised Regression with Instance-Varying Costs: Evaluation using Impact Curves}

\begin{document} 

\twocolumn[
\icmltitle{Binarised Regression with Instance-Varying Costs: Evaluation using Impact Curves}



\begin{icmlauthorlist}
\icmlauthor{Matthew Dirks}{}
\icmlauthor{David Poole}{}
\end{icmlauthorlist}


\begin{center}University of British Columbia (UBC), Vancouver, Canada\end{center}

\icmlkeywords{roc analysis, cost sensitive learning, model evaluation, model selection, cost or reward metrics, classification accuracy, utility, binary sorting, binary classification, application to mining, healthcare, entertainment, instance-varying costs, example-specific costs, case-conditional error costs, decision algorithms, prediction algorithms, cost curves, binarised regression, machine learning, ICML}

\vskip 0.3in
]
\begin{abstract} 
%
%
%
Many evaluation methods exist, each for a particular prediction task,
and there are a number of prediction tasks commonly performed including classification and regression. 
In binarised regression, binary decisions are generated from a learned regression model (or real-valued dependent variable), which is useful when the division between instances that should be predicted positive or negative depends on the utility. For example, in mining, the boundary between a valuable rock and a waste rock depends on the market price of various metals, which varies with time.
This paper proposes \emph{impact curves} to evaluate binarised regression with instance-varying costs, where some instances are much worse to be classified as positive (or negative) than other instances; e.g., it is much worse to throw away a high-grade gold rock than a medium-grade copper-ore rock, even if the mine wishes to keep both because both are profitable.
We show how to construct an impact curve for a variety of domains, including examples from healthcare, mining, and entertainment. 
Impact curves  
optimize binary decisions across all utilities of the chosen utility function,
identify the conditions where one model may be favoured over another,
and quantitatively assess improvement between competing models.
\end{abstract} 

\section{Introduction}
\label{sec:intro}
There is a long history of techniques for evaluating \textbf{models}
(e.g. classification or regression models, other prediction algorithms, or a noisy sensor reading)
in machine learning literature. 
The most prominent of these are for cases where there is a binary \textbf{target} 
(also known as ground truth, analyte, regressor, independent variable, or explanatory variable)
and binary \textbf{prediction}
(also known as dependent variable, outcome variable, or regressand)
placing instances into one of two categories (e.g. positive and negative).
The error of such problems can be measured in terms the \textbf{operating context};
in this case, the cost of false positives and false negatives. 
A utility function maps operating contexts (also known as outcomes or environment states) to a real-valued number which we want to maximize.

However, there are many cases where a binary decision needs to be made, but
where the target variable is more complicated than a binary true and false. 
Consider the following two examples:
\begin{itemize}
	\item In mining of rocks to extract minerals, we need
	to decide which rocks to process and which rocks to leave in the
	ground. There are percentages of various minerals in the rock. 
	The economics depends on the price of metals and the cost of processing, 
	both of which can change. 
	We would like to compare models across
	various operating contexts (various prices of metals and processing costs). 
	Some models might be better when the price of
	minerals is low and others might be better when the price is high. 

	\item Suppose, in a medical situation, we need to decide
	whether to go to the doctor or not, and there are different severities
	of going and not going to the doctor, depending on, for example, the
	level of cholesterol in the blood. Different people have different
	utilities, so we need to compare models across all
	utilities. Perhaps one model is better for
	hypochondriacs and those who are reluctant to go to the doctor. It is
	even possible that one model is best on average, but is dominated
	by some other model for each utility. 
\end{itemize}
In both of these cases, we would like to be able to see which
models are better and by how much, across various utilities.
In this paper we are concerned with binarised regression with instance-varying costs,
where a real-valued prediction is thresholded to produce a binary decision
for each instance, and each instance has its own cost associated it.
We develop a technique, which we call impact curves, specifically for this task.
The assumptions behind impact curves are as follows:

\begin{itemize}
	\item We have continuous predictions for all instances.
	These are often given by a learned regression model but could instead be a noisy sensor reading
	or other estimate.
	The units of the prediction do not have to match the target variable, because of the next point:

	\item A binary decision is made for each instance by thresholding its continuous prediction,
	such that we accept an instance if the predicted value is greater than the threshold 
	and reject it otherwise.

	\item Each instance has an associated cost or benefit when accepted and when rejected.

	\item We assume a linear utility function parameterized by $\theta$.
	The linear function should be chosen such that $\theta$ is a meaningful 
	measure of cost or profit in the application's domain. 
	It is reasonable to have a method that only compares
	utilities in a certain family of utility functions, as it is
	impossible to compare across \emph{all} utility functions as there are too many
	possible utility functions on real numbers (Section \ref{sec:taxonomyOfPredictionTasks} identifies common utility functions).

	\item We assume that the utility is additive across instances\footnote{This may
	seem like an innocuous assumption that is often made implicitly. It
	is often not true. For example in the mining scenario we may need
	enough rocks to fill a ship every day. In the medical example, as
	the number of patients increases so does the waiting time, until
	another clinic opens. The same problem happens if we report the
	average. We do not deal with this, but want to make the
	assumption explicit.}. 

	\item Given a utility function, we would then choose the threshold that maximizes utility.

\end{itemize}

\begin{figure}[htb]
	\begin{center}
	\centerline{\includegraphics[width=\columnwidth]{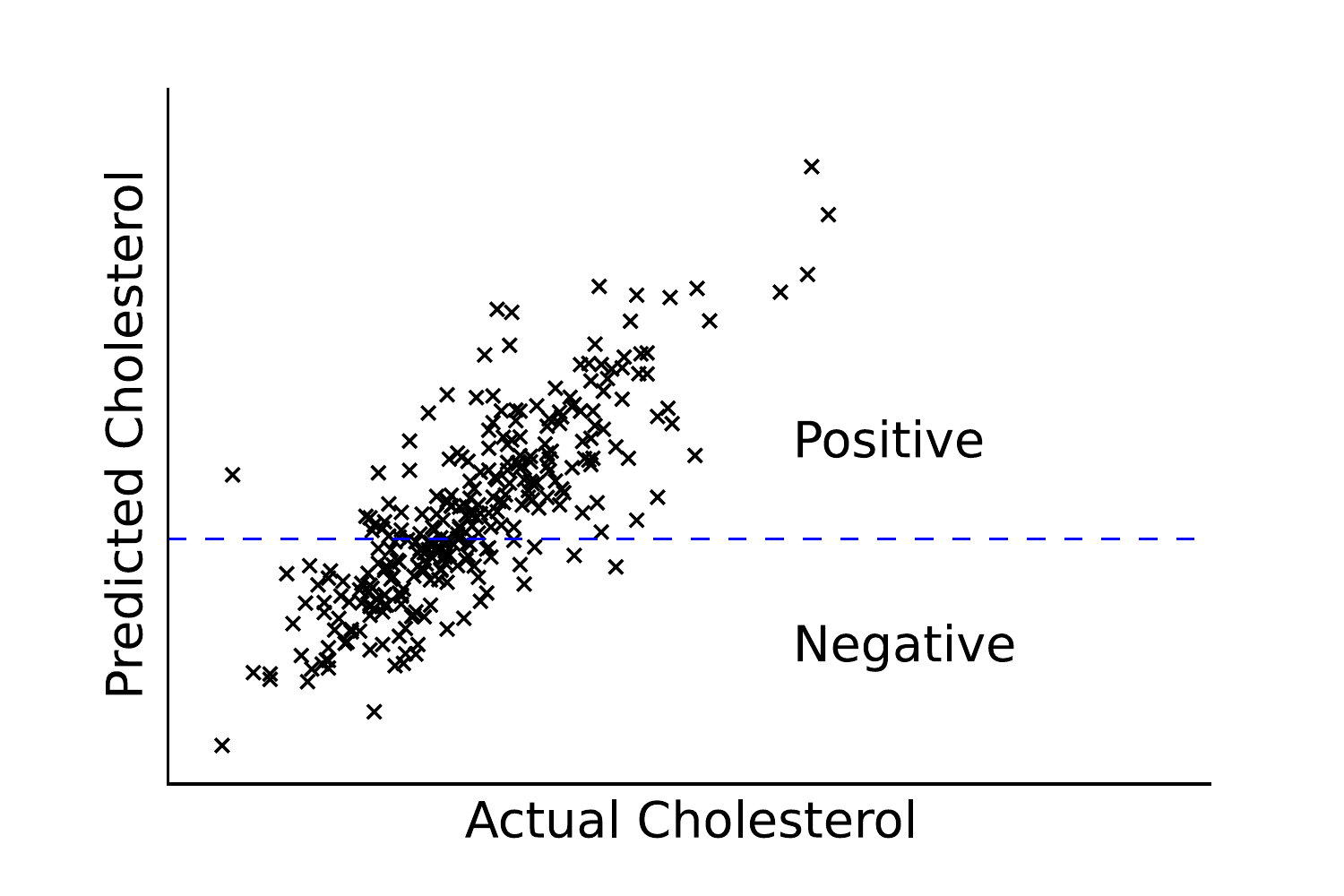}}
	\caption{Predicted cholesterol versus true cholesterol. The horizontal line is one possible threshold.}
	\label{fig:cholesterol_accept_reject}
	\end{center}
\end{figure}

For example, in the medical example,
the higher the cholesterol, the more important it is to go to the doctor. 
The utility function depends on the cost of visiting the doctor versus the benefit of taking care of your health. We let $\theta$ be the ratio between these two.
%
%
Fig. \ref{fig:cholesterol_accept_reject} shows a (learned) prediction of the
level of cholesterol versus the patient's actual cholesterol. 
We make a recommendation decision based on the
prediction: positive for each prediction
above the dashed line and negative for each instance below
the line. It is much more important to get the
decision for the extreme values correct than for those with a slightly
elevated cholesterol. 
The threshold used is chosen to maximize utility for each value of $\theta$. 
Section \ref{sec:example_chol} compares two different learned models. 

\begin{figure}[htb]
	\vskip 0.2in
	\begin{center}
	\centerline{\includegraphics[width=\columnwidth]{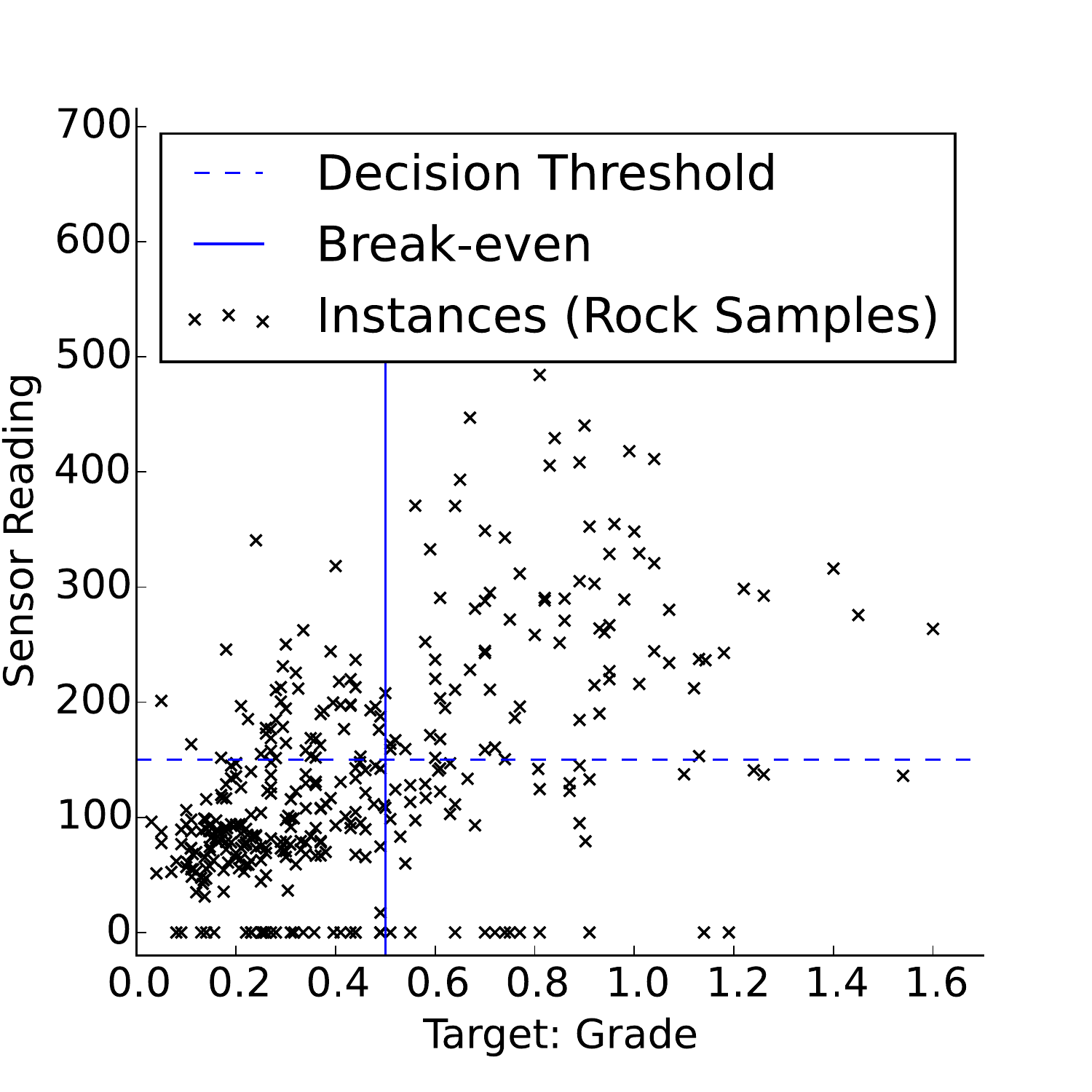}}
	\caption{Break-even point is for one possible utility (we explain more in Section \ref{sec:example_mining}). Instances (rock samples) above the threshold are accepted -- below are rejected.}
	\label{fig:mining_NA_predictions_AlgR1}
	\end{center}
	\vskip -0.2in
\end{figure}

In the rock mining example, the goal is to sort valuable ore from
waste rock. The cutoff between ``valuable'' and ``waste'' is the
break-even point when the tradeoff between processing costs and profit
from the rock are equal. The ratio of profit to processing costs is
unknown at the time of model building and evaluation, 
but it is more important to correctly classify the very high grade rocks
and the very low grade rocks than to correctly classify
a rock whose grade is hovering around the break-even point. 
In this example, the real-valued dependent variable is the output of a sensor
which can sense copper in a rock sample (data shown in
Fig. \ref{fig:mining_NA_predictions_AlgR1}). 
Impact curves let us compare the models across all possible break-even points, 
and to determine for which break-even points one model is preferred to another. 

In Section \ref{sec:taxonomyOfPredictionTasks} we identify related methods and the prediction tasks they apply to. Section \ref{sec:impactCurves} outlines the procedure to generate impact curves, and Section \ref{sec:examples} gives examples of applying impact curves to three domains.
\section{Related Work}
\subsection{Taxonomy of Misclassification Costs}
Costs come in a variety of forms, ranging from classification accuracy
to imbalanced costs for false negatives versus false positives
to instance-varying costs.
A taxonomy of the many forms of misclassification costs is in \cite{Turney2000TypesOfCost}.
Impact curves assume ``Conditional Error Costs,'' subcategorized as ``error cost conditional on feature value''.

\subsection{Taxonomy of Prediction Tasks}
\label{sec:taxonomyOfPredictionTasks}
The applicable model evaluation technique depends on the 
type of prediction task at hand and the utility.
Each technique assumes a particular family of utility functions.
%
A machine learning practitioner should be aware of the type of prediction
task they are trying to perform before performing evaluation. 
Known prediction tasks are outlined below and in Table \ref{tasksTable}.

\begin{table*}[tb]
\caption{Taxonomy of prediction tasks.}
\label{tasksTable}
\begin{center}
\begin{small}
\begin{sc}
\begin{tabular}{lccr}
\hline
\abovespace\belowspace
Prediction Task & target variable & predictions & utility is a function of \\
\hline
\abovespace
Classification & nominal or binary & nominal or binary & misclassifications \\
\hline
Classification with    & nominal or binary & nominal or binary & misclassifications and \\
Instance-Varying Costs &                   &                   & instance costs \\
\hline
Regression & continuous & continuous & deviation \\
\hline
Regression with        & continuous & continuous & deviation and instance costs \\
Instance-Varying Costs &            &            & \\
\hline
Ordinal Regression & ordinal & ordinal & deviation and rank costs \\
\hline
Binarised Regression & continuous & binary & misclassifications \\
\hline
Binarised Regression with & continuous & binary & misclassifications and \\
Instance-Varying Costs    &            &        & instance costs \\
\hline
\end{tabular}
\end{sc}
\end{small}
\end{center}
\end{table*}

\subsubsection{Classification}
\label{sec:classificationTask}
Classification is where 
the target variable and predictions are nominal (often binary)
for every instance.
ROC, for instance, assumes a binary target with binary predictions
and evaluates classification problems assuming a utility function over
	false negatives, 
	false positives,
	varying class distributions (a.k.a. class skew),
	and misclassification costs \cite{Metz1978_ROC,Majnik2013_ROCinML}.
There are also extensions to ROC for evaluating
multi-class classification \cite{RocSurvey}.

Precision-recall (PR) curves are an alternative to ROC when there exists a large skew in the class distribution \cite{Davis2006}; they are similar to ROC but lack a key feature of ROC which is that an algorithm that dominates in PR does not necessarily dominate across all utilities.

Cost curves are another alternative to the ROC representation \cite{costcurves},
providing improvements in the visualization and readability of performance \cite{drummond2004roc}. 
There is a strong relation between cost space and ROC space, including a point-to-point correspondence as shown in \cite{Orallo_ROC_vs_CC}.

Binary classifiers used for sorting can use an approach specifically designed to predict yield and quality called sorting optimization curves (SOC) \cite{Ooms2010}.
SOC has applications in the sorting of apples, seeds, processed food, and other items. Quality improvement and yield factors are used instead of the true positive and false positive rates as in ROC.

\subsubsection{Classification with Instance-Varying Costs}
Like classification, the target and predictions are nominal (or binary) in this task,
but utility is a function of per-instance costs
as well false positives (FP) and false negatives (FN).
ROC with instance-varying costs (ROCIV), developed by \cite{Fawcett2006ROCIV},
assumes binary classes.
In ROCIV graphs, the x-axis represents the fraction of total FP cost possible, and the y-axis represents the fraction of total TP benefits possible. 
Instances are essentially re-weighted prior to applying ROC methods.
Class distributions are assumed stable between training and deployment.


\subsubsection{Regression}
In regression the target and predictions are continuous
and utility is a function of the deviation between the prediction and the target variable, aggregated by summing or averaging.
Metrics for evaluating regression algorithms include
	mean absolute error (MAE, or mean absolute deviation),
	mean squared error (MSE),
	root mean squared error (RMSE),
	coefficient of determination ($r^2$),
	and others
	\cite{Ferri200927}.

Regression error characteristic (REC) curves generalize ROC curves to regression \cite{REC}.
REC curves plot error tolerance on the x-axis versus the percentage of points whose prediction is within the error tolerance on the y-axis.
REC assumes error in either direction is equally significant.

\subsubsection{Regression with Instance-Varying Costs}
Like regression, the target and predictions are continuous in this task,
but utility is a function of the costs and benefits of each instance
as well as the deviation between the prediction and target.
In utility-based regression, Torgo and Ribeiro
assign a relevance to each instance based on the particular domain \cite{UtilityBasedRegression2007}. 
Their example application is stock market forecasting
where the prediction of rare extreme values are most important 
(i.e. a sharp increase or decrease is more important than small variations).

\subsubsection{Ordinal Regression}
ROC analysis has been extended to the task of
ordinal regression \cite{RocOrdinalRegression},
where the target variable and predictions are ordered categories (ranks, e.g. movie ratings).
In this task utility is a function of the predicted and target rank of each instance.

\subsubsection{Binarised Regression}
In binarised regression the target and predictions are continuous,
but the prediction is thresholded at deployment time to produce a binary decision
(known as reframing).
%
Unlike regression, utility is function of false positives and false negatives
and evaluation can be performed across all utilities (i.e. misclassification costs vary).
This task is a hybrid of classification and regression.
The key differentia of this task is the volatile cutoff; the class of an instance is not known until deployment, at which point a threshold can be applied to the target and prediction.
Hernandez-Orallo et al. propose two solutions to the binarised-regression task and they break-down this task into further subcategories \cite{Orallo_BinarisedRegression_2016}.

\subsubsection{Binarised Regression with Instance-Varying Costs}
Consider the case when each instance has an associated cost
and a binary decision is made for each instance
but where the cutoff between positive and negative instances is not known until deployment.
Like binarised regression, 
the target variable and predictions are continuous and
predictions are reframed into binary classifications given a cutoff.
However, utility is a function of the continuous target variable as well as the binary misclassification rates.
%

\section{Impact Curves} 
\label{sec:impactCurves}
We propose impact curves to evaluate performance for binarised regression with instance-varying costs.
In this task, each model decides for each instance whether it is accepted
or rejected (or any other two classes of your choosing, such as positive or negative).


For simplicity we assume the utility function is parameterized by a single parameter, $\theta$.
%
Let the value of each instance be the difference between rejecting and accepting it,
such that the value of rejecting an instance is 0.
In other words, instead of incurring a cost for rejecting an instance, we give a reward for accepting it (or vice versa) -- 
this results in the same decisions being made.
The value of an instance, $v(i)$, has the following form:

\[
\label{eq:value_function}
  v(i,\theta) = \left.
  \begin{cases}
    f(i,\theta) & \text{if } accepted_i \\
    0 & \text{otherwise} \\
  \end{cases}
  \right.
\]
where $f(i,\theta)$ is determined by the domain and $accepted_i$ is whether or not instance $i$ was accepted. 
$f(i,\theta)$ is a linear function. 
Impact is the total value of all the instances:

\begin{equation}
\label{eq:impact}
I(\theta) = \sum_{i}{v(i,\theta)}
\end{equation}

where
$I(\theta)$ is impact,
$\theta$ is the operating context as defined in your domain,
$i$ is an instance,
and
$v(i,\theta)$ is the value of instance $i$ given $\theta$.
%
We assume the model has parameter(s) that can be tuned to change which instances it accepts (e.g. a threshold).

An impact curve is the upper envelope built from a model varied over its parametrizations. 
For example, picking the thresholds that maximize impact for each value of $\theta$ (see Fig. \ref{fig:mining_profitMaximization}).

Finally, we plot impact, $I(\theta)$, on the y-axis and operating context, $\theta$, on the x-axis.
Each model is piecewise linear in this space,
and each trivial decision is a line.
There are two trivial decisions:
the first accepts all instances,
and the second rejects all instances.
When comparing multiple models, a model is said to dominate if its impact is greater across all utilities.
Example impact curves are shown in Fig. \ref{fig:cholesterol_impact}, \ref{fig:mining_impact}, and \ref{fig:movies_impact}.

\section{Impact Curve Examples}
\label{sec:examples}

\subsection{Example: Mining}
\label{sec:example_mining}
When a mine sorts rock samples prior to processing and milling it is called pre-concentration,
which reduces costs and helps preserve the environment
because the amount of material sent to further downstream mining processes is
reduced 
\cite{Bamber_Klein_2008}.
%
%
The aim is to separate valuable high-grade rock samples from low-grade waste material.
A sample may be a single rock or may be a truckload of rocks.
A mine needs to know whether sorting at their site will increase profits,
and if so, 
which of many models to use
and how to optimally configure its parameters.

In this application sensors estimate the amount of copper in rock samples.
%
High-grade samples are valuable, called ore, and low-grade samples are waste.
The target is the grade of copper as given by geochemical assay for that rock.

\begin{figure}[bth]
\vskip 0.2in
\begin{center}
\centerline{\includegraphics[width=\columnwidth]{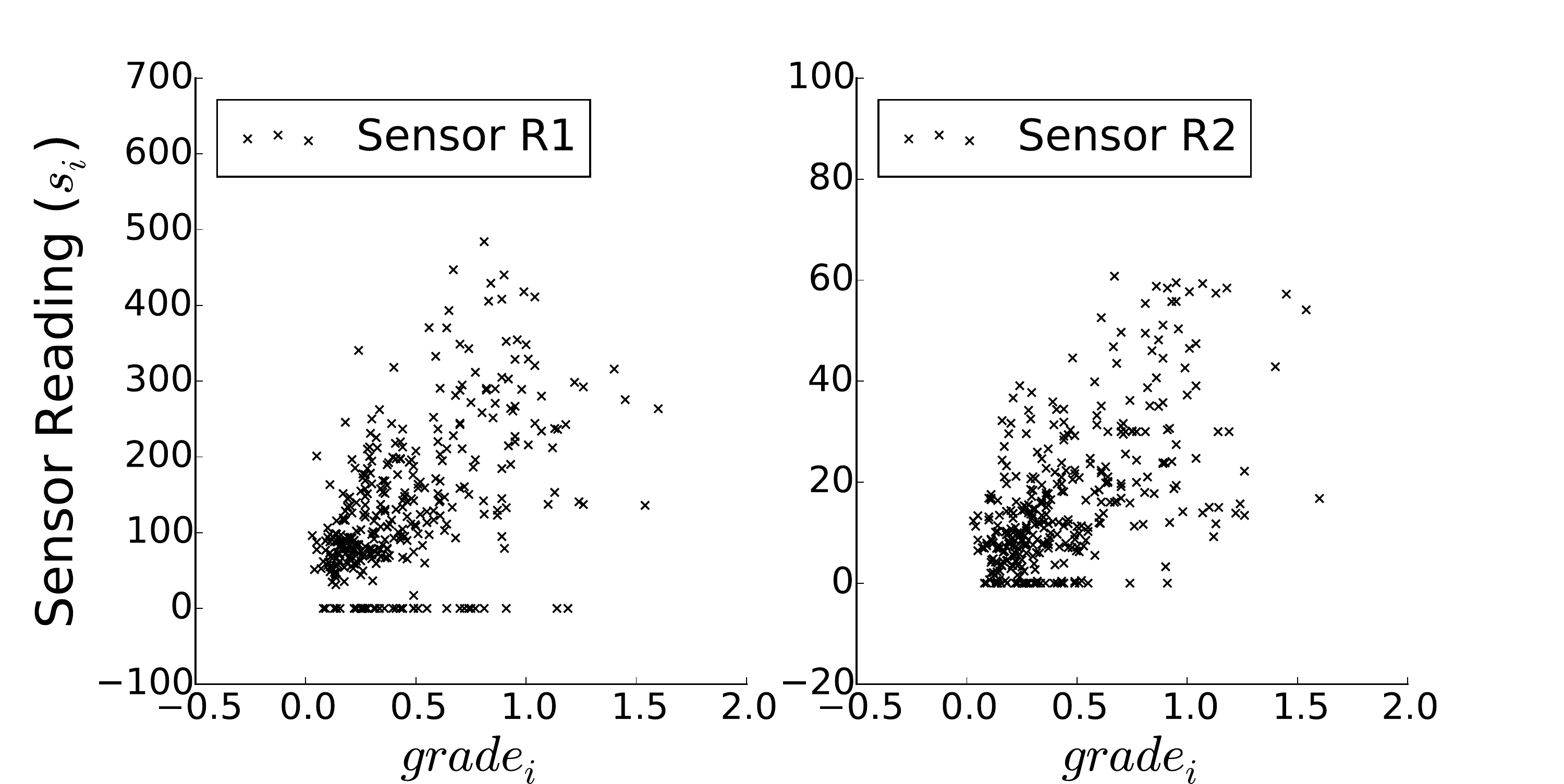}}
\caption{Rock metal sensor readings for two types of sensors.}
\label{fig:mining_NA_predictions}
\end{center}
\vskip -0.2in
\end{figure}

We have two sensors which produced readings shown in Fig. \ref{fig:mining_NA_predictions} for 363 rock samples.
The goal is to evaluate the performance of each for their ability to sort rocks into two bins: ore and waste. The sensor readings look similar and it isn't obvious which one is better. Traditional methods that measure prediction accuracy may tell us that one is better than another. It turns out that which is better depends on the price of copper and the processing costs. Impact curves will not only tell us this, but also at which point one is better than another.  

The cutoff (break-even point) between high-grade ore and low-grade waste depends on the mine's costs, the commodity being mined, and the price of that commodity.
Some of these factors may be confidential or unknown when doing an evaluation of a model.
%
%
%
Suppose the grade cutoff is 0.30\% Cu and we have two samples: 
one with 0.29\% Cu, and one with 0.10\% Cu. 
Both should be rejected, but maybe our model wrongly accepted both. 
The cost of accepting the 0.10\% Cu sample should be much higher than accepting the 0.29\% Cu sample 
because 0.29\% Cu is only slightly below the grade threshold whereas 0.10\% Cu is far from it.

The profit earned from rock $i$ is
\begin{equation}
\label{eq:rock_profit}
	grade_i \times mass \times price - cost
\end{equation}
where
$grade_i$ is the grade of the rock measured in percent copper (\%Cu),
$mass$ is assumed to be equal for all the rock samples so we ignore this constant for simplicity,
$price$ is the dollars per percent copper,
and
$cost$ is the cost of processing the rock.
We divide Eq. (\ref{eq:rock_profit}) by $cost$ and let $\theta$ subsume all the parameters because they don't depend on instance $i$:
\begin{align}
	\theta &= \frac{price}{cost} \\
	f(i,\theta) &= grade_i \times \theta - 1
\label{eq:rock_profit2}
\end{align}

\begin{figure*}[tbh]
	\begin{center}
	\centerline{\includegraphics[width=0.85\textwidth]{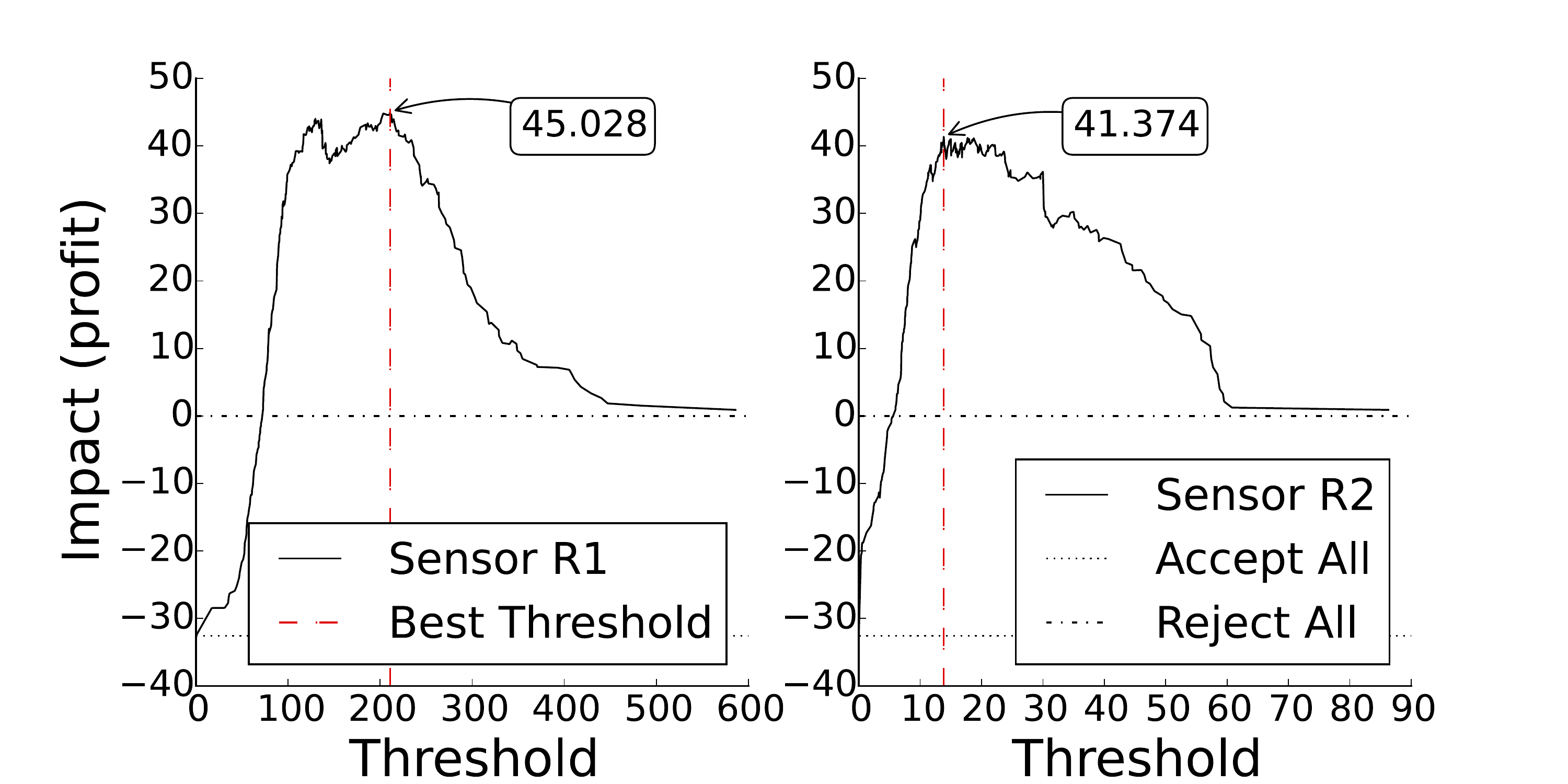}}
	\caption{Impact (or scaled profit) for sorting rock samples for a particular operating context: $\theta=2$. This is known as profit maximization. Note: we only have 362 rock samples, but the mine processes many more rock samples than this in a day, so profits will be multiplied considerably higher at production levels.}
	\label{fig:mining_profitMaximization}
	\end{center}
\end{figure*}

Now we are ready to form the value function as in Eq. (\ref{eq:value_function}):
\[
  v(i,\theta) = \left.
  \begin{cases}
    grade_i \times \theta - 1 & \text{if } s_i \geq threshold \\
    0 & \text{otherwise} \\
  \end{cases}
  \right.
\]
where
$i$ is an instance which is a sample of rock,
$\theta$ is the ratio of dollars per percent copper to processing costs,
$s_i$ is the copper sensor reading,
and $threshold$ is applied to the sensor readings producing a binary decision.
%
%
Note that Eq. (\ref{eq:rock_profit2}) is zero when the grade of a rock sample is equal to $\frac{cost}{price}$. Thus the inverse of $\theta$ can be thought of as a break-even point.
At the time of writing, the cost of 1\% copper is $8.16$ USD. 
For one mine, the processing cost for accepting a sample may be $4.08$ and the
cost for rejecting a sample is $0$.
Therefore, $\theta = \frac{8.16}{4.08-0} = 2$, and we break-even when sample grade is 0.5 \%Cu.
Using $\theta=2$, two models -- Sensor R1 and Sensor R2 -- are shown in Fig. \ref{fig:mining_profitMaximization}
with impact on the y-axis and all possible thresholds on the x-axis.
When the processing cost is 1\$ impact is profit in dollars and otherwise impact is simply a scaled version of profit (due to our simplification in Eq. (\ref{eq:rock_profit2})).
This step is essentially \textit{profit maximization}.
We use the term ``impact'' instead of profit because in general we may not be dealing with units of profit or dollars.

\begin{figure}[htb]
	\begin{center}
	\centerline{\includegraphics[width=0.85\columnwidth]{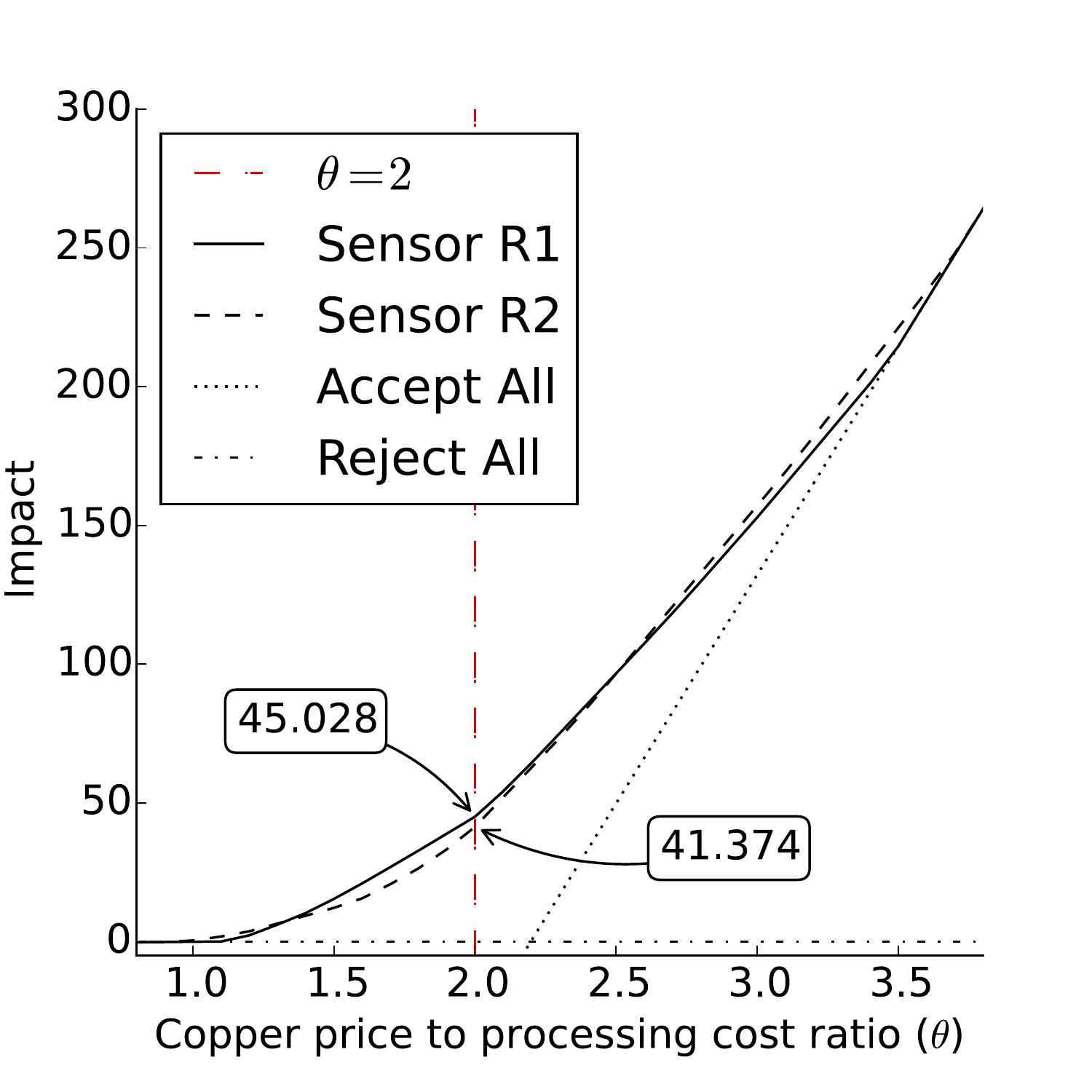}}
	\caption{Impact curve for sorting rock samples based on sensor readings.}
	\label{fig:mining_impact}
	\end{center}
\end{figure}

At $\theta=2$ we can see from Fig. \ref{fig:mining_profitMaximization} that the best threshold for Sensor R1 is 210 and 14 for Sensor R2. 
Impact curves maximize impact for each value of $\theta$, which can be seen in
the corresponding impact curves shown in Fig. \ref{fig:mining_impact}.
Fig. \ref{fig:mining_profitMaximization} and \ref{fig:mining_impact} show that R1 has the higher impact for $\theta=2$: 45.028 vs $41.374$.

\begin{figure}[htb]
	\begin{center}
	\centerline{\includegraphics[width=0.85\columnwidth]{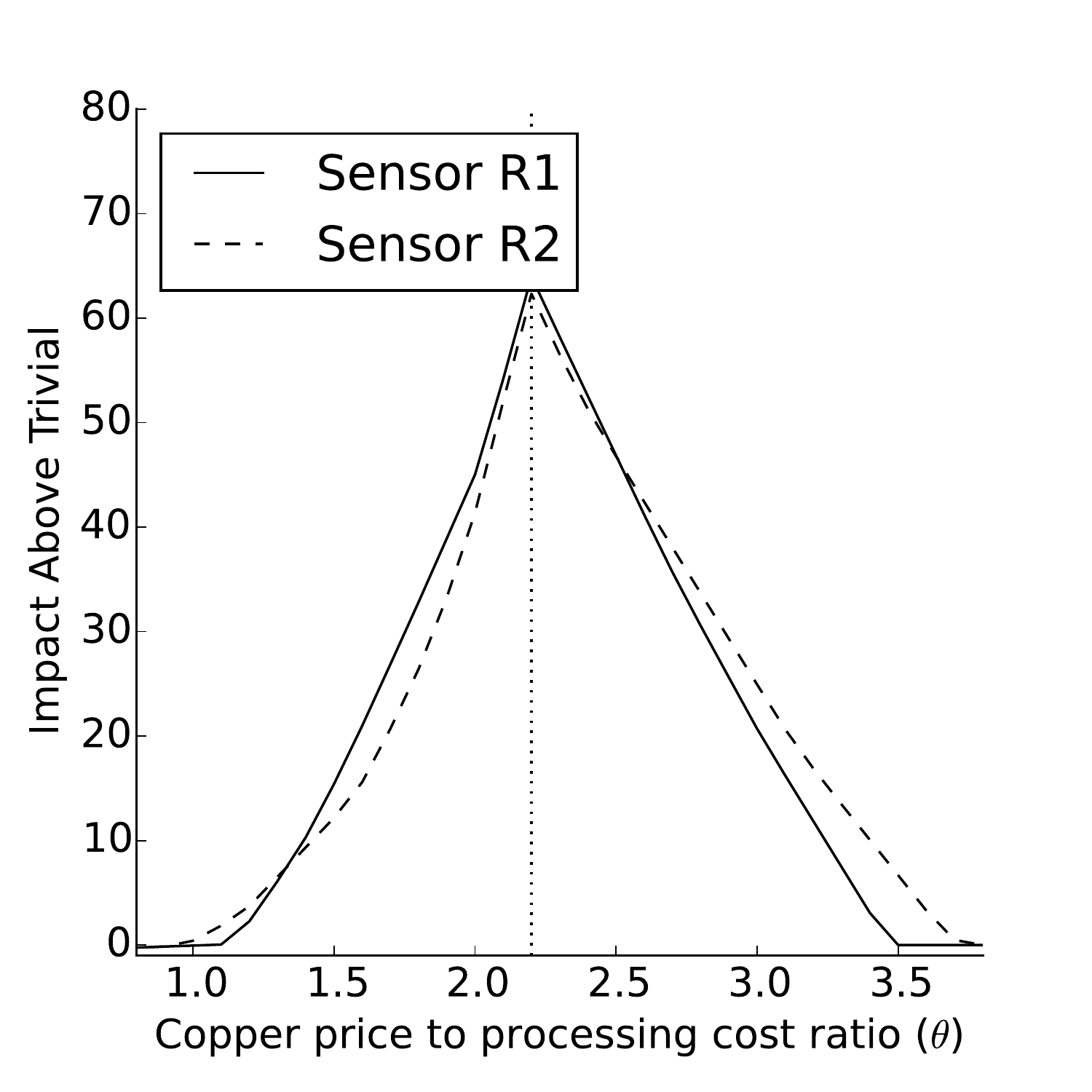}}
	\caption{Impact curve, relative to trivial impact, for sorting rock samples based on sensor readings.
	The vertical dotted line shows where the best trivial algorithm switches from ``reject all'' (left side) to ``accept all'' (right side).}
	\label{fig:mining_impact_improvement}
	\end{center}
\end{figure}

The two models may appear close, but even a small increase in profit has a large effect in a big mine.
Furthermore, we can take the difference between the impact of each algorithm and the trivial impact to visualize the improvement each algorithm is making, as shown in Fig. \ref{fig:mining_impact_improvement}.
Using impact curves or impact curve improvement we can choose the best model if given a value for $\theta$ (i.e. the price of copper and processing cost, or the break-even cutoff).
In Fig. \ref{fig:mining_impact} and \ref{fig:mining_impact_improvement} we see that R1 is better between $\theta=1.4$ to $2.5$.

\subsection{Example: Cholesterol \& Heart Attacks}
\label{sec:example_chol}
When your blood has too much bad cholesterol it builds up in the arteries, restricting blood flow to the heart. If not enough blood and oxygen can reach the heart, a heart attack occurs. 
In this example we use a heart disease dataset from \cite{UCI_ML_repository}%
\footnote{Data collected by Andras Janosi, William Steinbrunn, Matthias Pfisterer, and Robert Detrano. We used the WEKA version of the database downloaded from \url{http://prdownloads.sourceforge.net/weka/datasets-numeric.jar}, filename ``cholesterol.arff''.}
to predict a person's cholesterol level and decide if they should go see a doctor.
According to the National Institutes of Health%
\footnote{\url{https://www.nlm.nih.gov/medlineplus/magazine/issues/summer12/articles/summer12pg6-7.html}},
total cholesterol can be categorized into three risk levels:
less than 200 mg/dL is desirable, 200 - 239 mg/dL is borderline high, and greater than 240 mg/dL is high.

\begin{figure}[htb]
\begin{center}
\centerline{\includegraphics[width=\columnwidth]{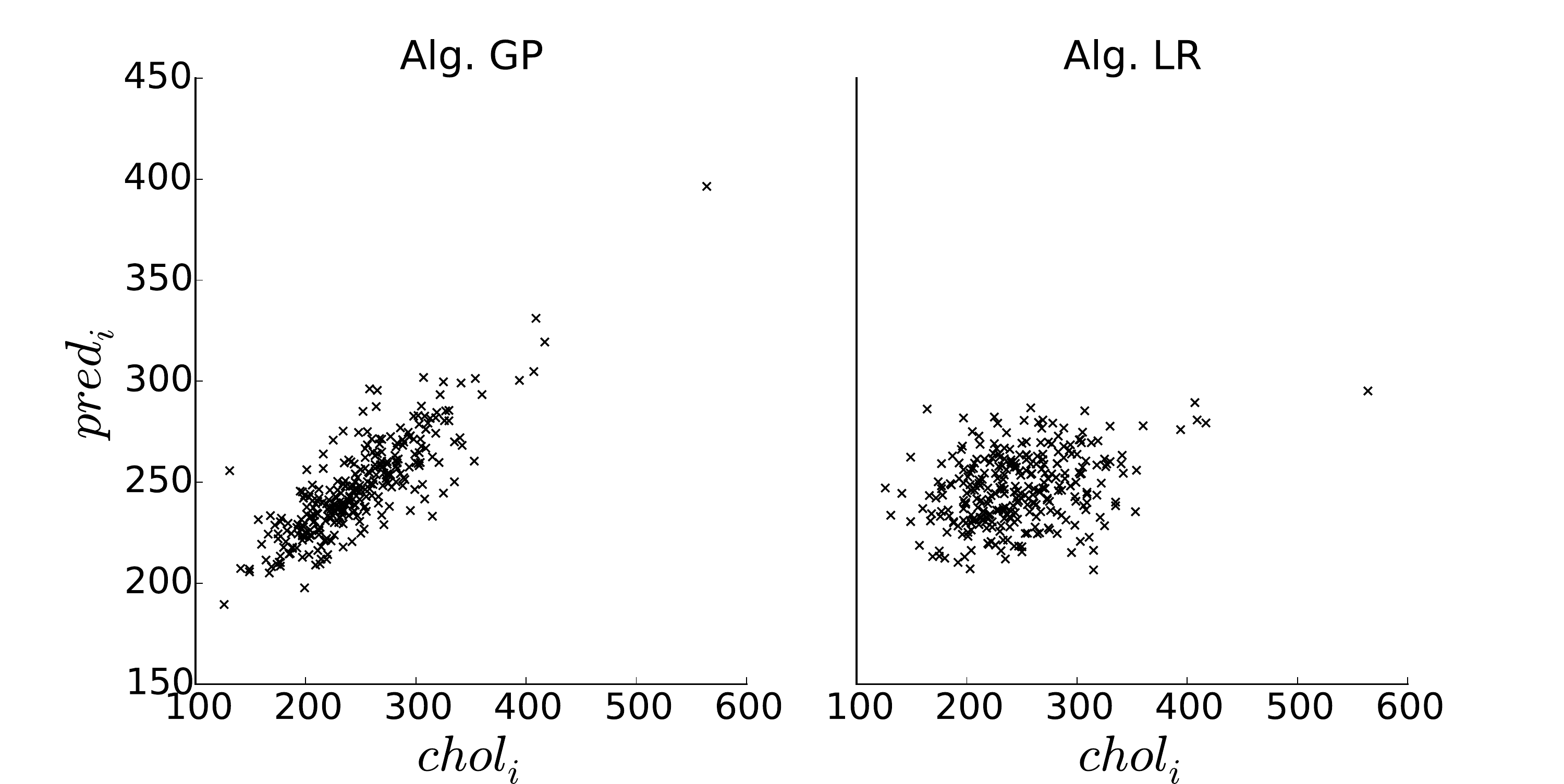}}
\caption{Cholesterol predictions.}
\label{fig:cholesterol_predictions}
\end{center}
\end{figure}

%
%
We trained two models on 303 examples: linear regression (Alg. LR) and a Gaussian process (Alg. GP) implemented in WEKA. Their predictions are shown in Figure \ref{fig:cholesterol_predictions}. We threshold the prediction to produce a binary decision.

\begin{figure}[htb]
\begin{center}
\centerline{\includegraphics[width=\columnwidth]{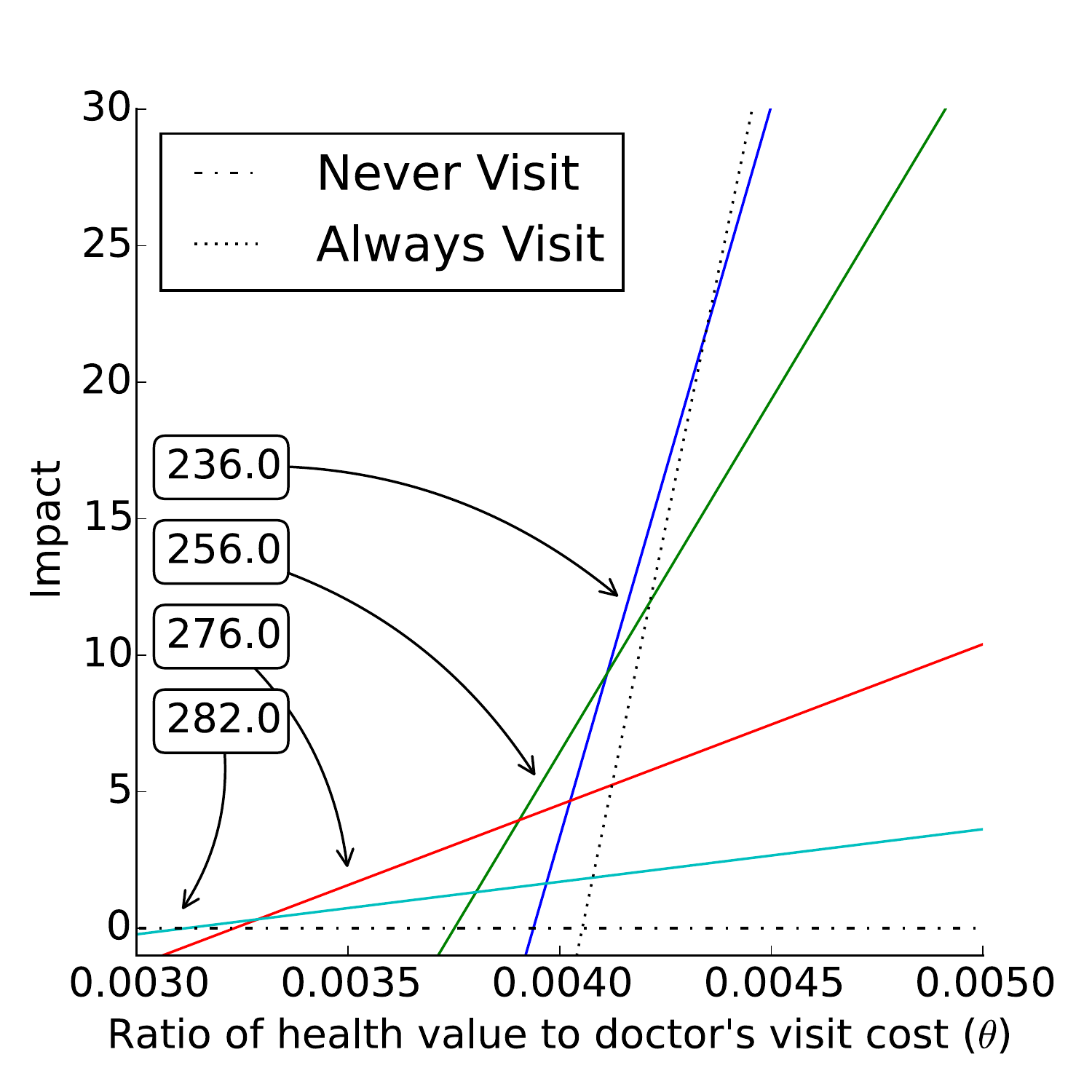}}
\caption{Impact lines for 4 thresholds on cholesterol predictions given by linear regression (Alg. LR). Only 4 thresholds are shown here for simplicity. $\theta$ is the ratio of health value to doctor's visit cost. The annotated arrows show the threshold value for each line.}
\label{fig:cholesterol_family}
\end{center}
\end{figure}

\begin{figure}[htb]
\begin{center}
\centerline{\includegraphics[width=\columnwidth]{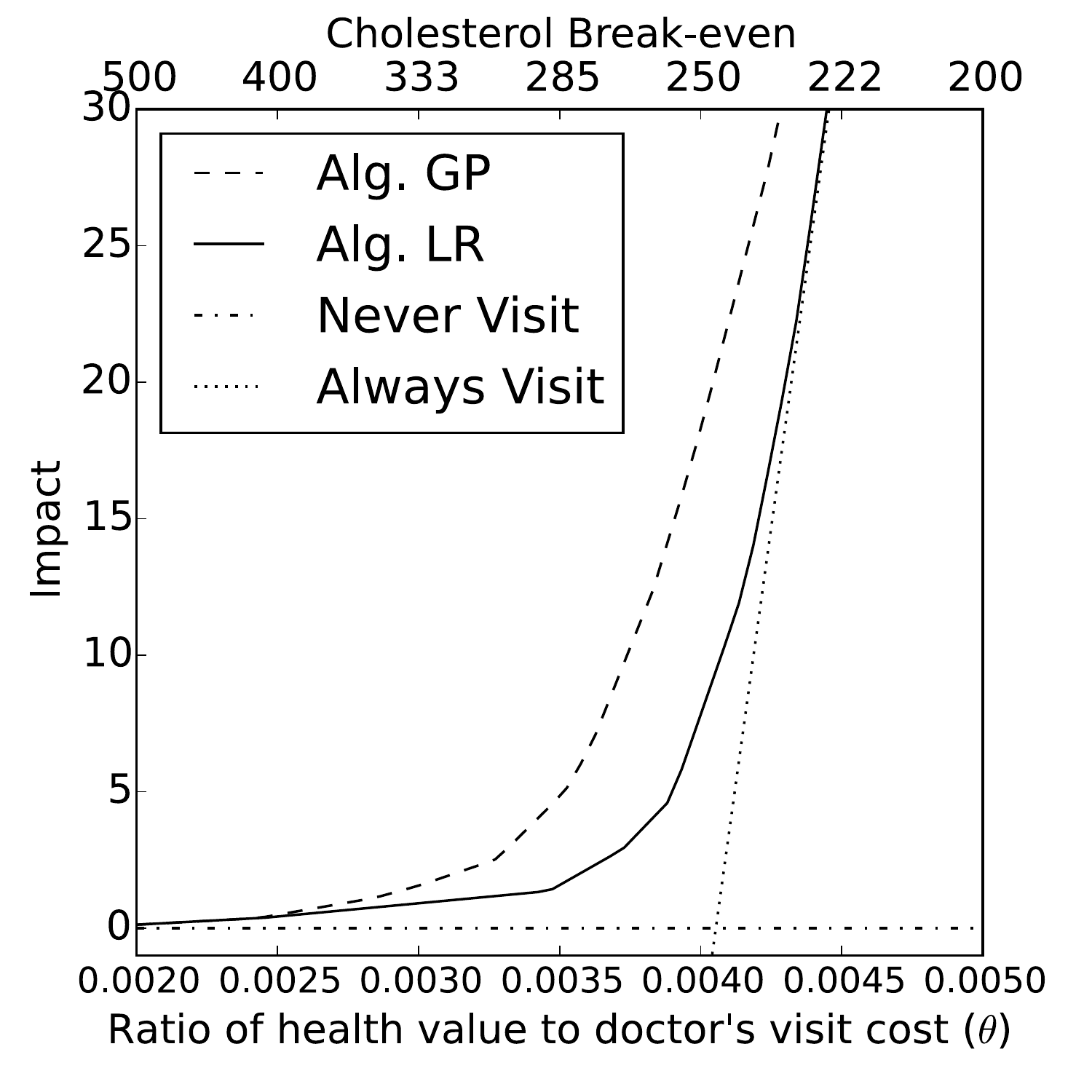}}
\caption{Impact curve for deciding whether to visit the doctor. $1/\theta$ is the break-even point at which it is equally valued to go to the doctor as not to go.}
\label{fig:cholesterol_impact}
\end{center}
\end{figure}

The value of visiting the doctor is $\theta$ times a person's cholesterol less the cost of visiting the doctor. Like in the mining example, we divide by the cost of visiting the doctor so that $\theta$ expresses the benefit of health relative to the cost of visiting the doctor. We define the value function as:

\[
  v(i,\theta) = \left.
  \begin{cases}
    chol_i \times \theta - 1 & \text{if } pred_i \geq threshold \\
    0 & \text{otherwise} \\
  \end{cases}
  \right.
\]

where
$v(i,\theta)$ is the value associated with person $i$,
$\theta$ is the ratio of health value to doctor's visit cost,
$chol_i$ is person $i$'s total cholesterol,
$pred_i$ is the prediction of person $i$'s total cholesterol given by a learned model,
and $threshold$ turns the prediction into a binary decision.
An example using a linear regression (Alg. LR) algorithm is shown in Fig. \ref{fig:cholesterol_family}. The upper envelope of these is an impact curve, shown as a solid black line in Fig. \ref{fig:cholesterol_impact}. A useful way to interpret $\theta$ is to consider the point at which it is equally valuable to predict positive as negative; this is the break-even point. In this example the break-even point is $1/\theta$ which is shown on the top x-axis of Fig. \ref{fig:cholesterol_impact}.

The impact curves for the two models are shown in Fig. \ref{fig:cholesterol_impact} and in this case Alg. GP dominates over Alg. LR because its impact is better for all utility (we do not claim any medical expertise -- please consult a doctor if you are making decisions about your cholesterol levels).

\subsection{Example: Movie Ratings}
\label{sec:example_movies}
In this example the task is to decide if a movie should be recommended to a user.
We want the user to be happy with our recommendation.
We assume that a high rating indicates they were happy with the movie,
but we don't know exactly how high is necessary -- we just know that higher is better.

We build a model by predicting the rating a user will give to a movie,
then threshold the prediction into two classes: recommend and don't recommend.
The predicted ratings are in $\mathbb{R}$, whereas the true ratings (target variable) are in $\mathbb{N}$ (specifically 1, 2, 3, 4, and 5 stars).

We trained two prediction algorithms%
\footnote{Prediction algorithms were two configurations of a Python implementation of relational collaborative filtering, available at \url{http://artint.info/code/python/}}
on the movielens dataset%
\footnote{Movie ratings dataset downloaded from \url{http://files.grouplens.org/datasets/movielens/ml-100k/u.data}}
consisting of many users' rating of many movies.
We call these two algorithms Alg. M1 and Alg. M2.

The value of recommending a movie depends on the cutoff between happy and unhappy,
we could guess and say that 3 stars or more is when users are happy,
but instead we use the impact curve technique to evaluate the performance for all possible cutoffs.
Therefore, the value of recommending a movie that makes the user happy is the difference between the user's rating of the movie and the cutoff:
\[
\label{eq:impact_movie}
  v(i,\theta) = \left.
  \begin{cases}
    rating_i - \theta & \text{if } pred_i \geq threshold \\
    0 & \text{otherwise} \\
  \end{cases}
  \right.
\]

where
$v(i)$ is the value associated with user-movie pair $i$,
$rating_i$ is the true rating,
$\theta$ is the cutoff between happiness and unhappiness (it is a parameter of the utility function),
$pred_i$ is the predicted rating given by a prediction algorithm,
and $threshold$ is the cutoff applied to the prediction.

\begin{figure}[bth]
\begin{center}
\centerline{\includegraphics[width=\columnwidth]{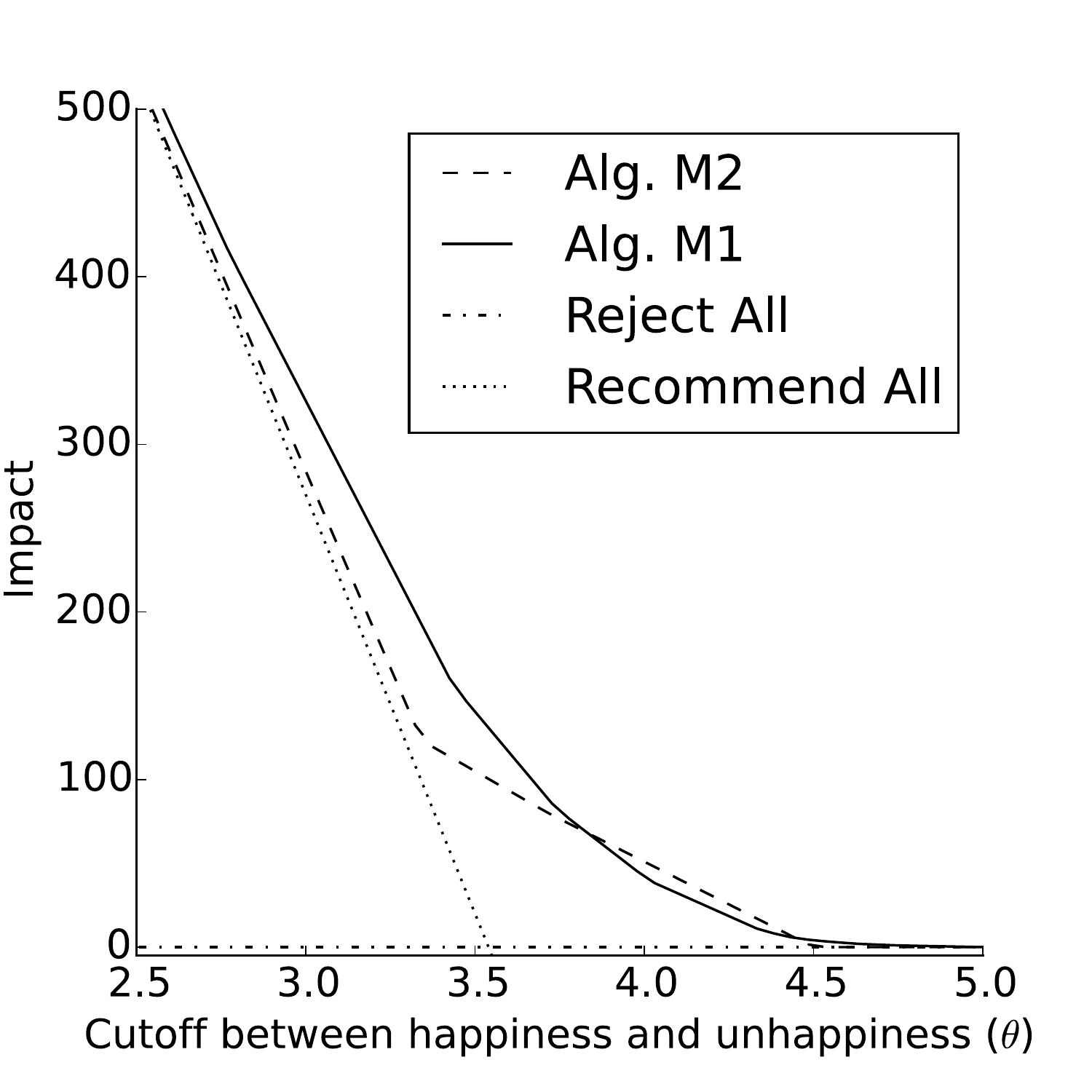}}
\caption{Impact curve for movie recommender.}
\label{fig:movies_impact}
\end{center}
\end{figure}

The impact curves for the two models are shown in Fig. \ref{fig:movies_impact}. For $\theta < 3.7$ Alg. M1 should be used and otherwise Alg. M2 should be selected.

\section{Limitations and Future Work}
	Impact curves allow evaluation and model selection to be performed
	on a training dataset.
	Cross-validation should also be considered.
	%
	%
	This technique should therefore be considered as one that chooses the optimal threshold in addition to model selection and performance evaluation. But care has to be taken since we cannot ensure that the optimal threshold at training time will be the optimal threshold at deployment time.





Impact curves limit the parametrization of utility to only one parameter. 
Using one parameter allowed the visualization of impact in 2D.
Future work is to visualize impact curves when dimensionality increases.

Impact curves assume binary decisions.
This could be extended to multi-class (categorical) decisions.
Future work is to extend impact curves to non-regression tasks with binary decisions.
The utility function is assumed to be linear. Future work is to relax this assumption.

Summing instance values may not be applicable to some domains, future work is to generalize to cover cases where the total impact may depend on interactions between instances.

\section*{Conclusion}
We have identified the task of instance-varying costs within binarised regression, where utility is unknown at training time and a binary decision is made for each instance.
We have proposed a method, called impact curves, specifically for this task.
Impact curves are used to perform model selection and evaluation.
We developed one possible visualization: with $impact$ on the y-axis versus operating contexts on the x-axis.
%
%
Finally, we applied the impact curve technique to three applications -- heart attack prevention, mineral sorting, and movie recommendations -- and showed how to create impact curves for each.

\matt{8 pages for content max. Pages 9 and 10 can have references/acknowledgements.}

\nocite{Boutilier2003}
\FloatBarrier



\section*{Acknowledgements} 
\iffalse
	\input{tex_files/ack.tex}
\else
	(Removed for blind review)
\fi


\bibliography{paper}
\bibliographystyle{icml2017}

\end{document}